\documentclass[10pt]{article}

\usepackage[margin=1in]{geometry}
\usepackage{amsmath,amssymb}
\usepackage{graphicx}
\usepackage{booktabs}
\usepackage[protrusion=true,expansion=false]{microtype}
\usepackage[numbers,sort&compress]{natbib}
\usepackage{xcolor}
\usepackage[colorlinks=true,linkcolor=blue!60!black,citecolor=blue!60!black,urlcolor=blue!60!black]{hyperref}
\usepackage{caption}
\newcommand{\BiasAgEnsHi}{0.441}

\newcommand{\BiasAgHeckHiNoInstr}{0.582}

\newcommand{\BiasAgHeckTSHi}{0.002}
\newcommand{\BiasAgHeckTSHiNoInstr}{0.778}

\newcommand{\BiasAgIWHi}{0.678}

\newcommand{\CovAgBlindHi}{69.3\%}

\newcommand{\CovAgEnsHi}{64.4\%}
\newcommand{\CovAgEnsHiNoInstr}{28.8\%}

\newcommand{\CovAgEnsMid}{77.8\%}

\newcommand{\CovAgHeckHi}{89.1\%}
\newcommand{\CovAgHeckHiNoInstr}{84.7\%}

\newcommand{\CovAgHeckTSHi}{88.9\%}
\newcommand{\CovAgHeckTSHiNoInstr}{40.3\%}

\newcommand{\CovAgHeckTSMid}{88.8\%}

\newcommand{\CovAgIWHi}{43.1\%}

\newcommand{\ECEAgEnsHi}{0.266}

\newcommand{\ECEAgHeckTSHi}{0.036}

\newcommand{\ECEAgOracleHi}{0.012}
\newcommand{\EOneSelFrac}{50\%}
\newcommand{\EZeroDevMLE}{$5.0\times10^{-7}$}
\newcommand{\EZeroDevTS}{$2.8\times10^{-6}$}
\newcommand{\EZeroLLDev}{0.000}
\newcommand{\EZeroN}{5574}
\newcommand{\EZeroNSel}{4281}
\newcommand{\EZeroRhoMLE}{0.7356}

\newcommand{\EZeroSigmaMLE}{1.5701}

\newcommand{\PKillIWvsHeck}{0.008}

\newcommand{\RealCovCalHeckTS}{77.5\%}

\newcommand{\RealCovWellCalDrop}{93.2\%}

\newcommand{\RealCovWellCalGP}{98.0\%}

\newcommand{\RealCovWineHeckTS}{64.1\%}

\newcommand{\RealCovZeroCalEns}{73.9\%}
\newcommand{\RealCovZeroCalHeckTS}{71.8\%}

\newcommand{\RealECECalDrop}{0.175}
\newcommand{\RealECECalEns}{0.201}
\newcommand{\RealECECalGP}{0.114}

\newcommand{\RealECECalHeckTS}{0.051}
\newcommand{\RealECECalIW}{0.235}
\newcommand{\RealECECalSky}{0.026}

\newcommand{\RealECEWineDrop}{0.200}
\newcommand{\RealECEWineEns}{0.223}
\newcommand{\RealECEWineGP}{0.164}

\newcommand{\RealECEWineHeckTS}{0.138}
\newcommand{\RealECEWineIW}{0.252}
\newcommand{\RealECEWineSky}{0.049}
\newcommand{\RhoHatHi}{0.90}
\newcommand{\RhoHatHiNoInstr}{0.14}
\newcommand{\RhoHatHiSD}{0.02}

\newcommand{\VigImageNetReaLRho}{0.87}
\newcommand{\VigImageNetReaLSE}{0.09}
\newcommand{\VigMTRho}{-0.90}
\newcommand{\VigNBoundary}{8}
\newcommand{\VigNTotal}{13}

\newcommand{\E}{\mathbb{E}}

\newcommand{\IMR}{\lambda}

\title{Heckman-Corrected Epistemic Uncertainty:\\
Selection on Unobservables Defeats Importance Weighting}

\author{Gunner Levi Howe\\
Independent Researcher\\
\texttt{gunnerlevihowe@gmail.com}}

\date{July 2026}

\begin{document}
\maketitle

\begin{abstract}
Training data for machine-learning models is routinely collected by a
selection process the model never sees: loans are observed only when
granted, outcomes only when a test was ordered, regions of input space
sampled only when someone chose to sample them. The field's default fixes
--- importance weighting, covariate-shift correction, MAR imputation ---
assume selection is ignorable given observables. Econometrics solved the
harder problem in 1979: Heckman's two-equation model jointly fits a probit
selection equation and an outcome equation linked through correlated
errors, and the inverse-Mills-ratio term corrects the outcome model for
selection on \emph{unobservables} --- the case where importance weighting
is structurally helpless. We instantiate that machinery for deep epistemic
uncertainty: a deep outcome network, a linear selection head, and a joint
bivariate-normal likelihood over all units (observed and not), ensembled
for epistemic variance. Our implementations first reproduce the
seven-digit Stata reference output for both the classic two-step and the
maximum-likelihood estimator on the RAND Health Insurance data (max
coefficient deviation \EZeroDevTS{} two-step, \EZeroDevMLE{} MLE). In a
controlled generator where sampling probability depends on an unobservable
correlated ($\rho$ swept $0\to0.9$) with the outcome noise, deep
ensembles, MC dropout, and GP baselines are overconfident exactly where
data was avoided for reasons correlated with the outcome (coverage of
nominal-90\% intervals falls to \CovAgEnsHi{} at $\rho=0.9$), and
importance weighting with \emph{oracle} propensities does not fix it
(\CovAgIWHi{}) --- reweighting corrects the covariate distribution, not
the conditional bias $\E[y \mid x, s{=}1] \neq \E[y \mid x]$. The
Heckman-corrected predictive distribution restores coverage
(\CovAgHeckTSHi{}) when the selection equation contains an
\emph{instrument} --- a variable affecting selection but not the outcome
--- and degrades gracefully but measurably without one (\CovAgHeckTSHiNoInstr{}); we
sweep this honesty curve rather than hide it. The joint MLE needs a
warm-up schedule to match the two-step's stability with deep feature maps
(otherwise the flexible outcome net absorbs the correction), a methods
finding we characterize. On real tabular data with induced, documented MNAR
selection, the corrected intervals are the best-calibrated (lowest
region-ECE) of every non-oracle method in selected-against regions on both
datasets; baselines that match its raw coverage do so only by over-widening
everywhere, which region-ECE exposes. A discussion vignette fits the same
two-equation model to public benchmark reporting panels (Papers-with-Code
tables), where the no-instrument pathology our controlled experiments
quantify appears in the wild: most fits hit the $|\hat\rho|=1$ boundary.
We state plainly which identification regime a practitioner is in, and
release the faithfulness-gated implementation.
\end{abstract}

\section{Introduction}
\label{sec:intro}

Every published treatment of distribution shift in uncertainty
quantification (UQ) that we are aware of corrects for selection on
\emph{observables}: importance weighting and covariate-shift adaptation
reweight by a propensity that depends on $x$
\citep{shimodaira2000,sugiyama2007,cortes2008}, and evaluations of
predictive uncertainty under shift perturb the marginal $p(x)$
\citep{ovadia2019}. The harder and, in deployed systems, common case is
selection on \emph{unobservables}: the probability that a unit enters the
training set depends on latent determinants of the outcome itself. A bank
observes repayment only for granted loans, and the loan officer's
side-information correlates with default risk; a hospital records disease
severity only for admitted patients; a simulation campaign samples
parameter regions an expert already believed well-behaved. In all these
cases $\E[y \mid x, s{=}1] \neq \E[y \mid x]$, so \emph{no reweighting of
the selected sample --- even with exact propensities --- can recover the
population regression}. A model trained on such data is not merely
data-poor in the selected-against regions; it is systematically wrong
there, and an epistemic-uncertainty method that widens intervals with
data sparsity alone will be confidently wrong.

Econometrics has owned this problem since \citet{heckman1979} (building
on \citealp{tobin1958,heckman1974}): model selection and outcome jointly,
\begin{align}
s_i &= \mathbf{1}\!\left[w_i^\top\gamma + u_i > 0\right], &
y_i &= f(x_i) + \varepsilon_i \quad \text{observed iff } s_i = 1,
\label{eq:heckman}
\end{align}
with $(u_i, \varepsilon_i)$ bivariate normal, $\mathrm{Corr}(u_i,
\varepsilon_i) = \rho$. The selected-sample regression then satisfies
$\E[y \mid x, w, s{=}1] = f(x) + \rho\sigma\,\IMR(w^\top\gamma)$ where
$\IMR = \phi/\Phi$ is the inverse Mills ratio: the bias term is an
explicit, estimable function, and the population $f(x)$ is recoverable.
This paper instantiates that machinery with deep feature maps and asks
whether it repairs epistemic uncertainty where importance weighting
cannot.

\paragraph{The identification caveat, up front.} Heckman identification
is robust when the selection equation contains an \emph{exclusion
restriction} --- an instrument $z$ that affects selection but not the
outcome. Without one, identification rides entirely on the bivariate-normal
functional form and is notoriously fragile
\citep{puhani2000,vella1998,winship1992}. Every experimental grid in this
paper therefore contains instrument-present \emph{and} instrument-absent
conditions, and we report the degradation curve rather than the best
case. In ML settings natural instruments exist and should be named:
which sensor, site, annotator, or batch collected the datum;
acquisition-policy randomness such as an exploration epsilon; rung-
assignment randomness in successive halving (exploited in the companion
paper \citep{howe2026survivor}). A practitioner without any such variable
should read our no-instrument curves as their expected reality.

\paragraph{Contributions.}
\begin{enumerate}
\item \textbf{Faithfulness-gated implementation} (\S\ref{sec:e0}): probit,
two-step, and joint-MLE Heckman estimators that reproduce the seven-digit
Stata reference output of \citet{cameron2005} on the RAND HIE data
(max deviation \EZeroDevTS{} / \EZeroDevMLE{}) and statsmodels' probit to
$10^{-8}$; pre-registered as a gate before any experiment.
\item \textbf{Deep Heckman UQ} (\S\ref{sec:method}): a deep outcome net
with a linear selection head trained by the joint bivariate-normal
likelihood over all units, plus a two-step variant (probit, then an
inverse-Mills feature); ensembles of either give a selection-corrected
predictive distribution.
\item \textbf{The controlled demonstration} (\S\ref{sec:e1}): under
selection on unobservables, oracle-propensity importance weighting fails
where the Heckman correction succeeds --- coverage \CovAgIWHi{} vs
\CovAgHeckTSHi{} at $\rho = 0.9$ (nominal 90\%) in selected-against
regions --- and the pre-registered kill condition (IW matching Heckman
would falsify the premise) does not fire. Without the instrument the
correction degrades to \CovAgHeckTSHiNoInstr{}; we quantify the honesty
curve in $\rho$.
\item \textbf{A methods finding} (\S\ref{sec:e1}): with deep feature maps
the two-step estimator is markedly more stable than the joint MLE, which
requires a warm-up schedule to avoid absorbing the correction term into
the flexible outcome function.
\item \textbf{Semi-real validation and a real-world vignette}
(\S\ref{sec:e2}--\ref{sec:e4}): induced, fully documented MNAR selection
on real tabular tasks where the corrected intervals are the
best-calibrated (lowest region-ECE) non-oracle method in selected-against
regions, and where we show raw coverage misleads; and a Heckman analysis
of public benchmark reporting panels in which the no-instrument boundary
pathology ($|\hat\rho| = 1$ in \VigNBoundary{} of \VigNTotal{} fits)
appears exactly as the controlled experiments predict.
\end{enumerate}

\section{Background and related work}
\label{sec:related}

\paragraph{Heckman selection models.} The two-step estimator
\citep{heckman1979} fits a probit for $P(s{=}1 \mid w) = \Phi(w^\top
\gamma)$, then regresses $y$ on $[x, \IMR(w^\top\hat\gamma)]$ over the
selected sample; $\sigma$ and $\rho$ are recovered from the residual
variance and the Mills coefficient $\beta_\IMR = \rho\sigma$. The joint
MLE maximizes the bivariate-normal likelihood
\begin{equation}
\ell = \sum_{s_i=0}\log\Phi(-w_i^\top\gamma)
+ \sum_{s_i=1}\left[\log\phi_\sigma(y_i - f(x_i)) +
\log\Phi\!\left(\tfrac{w_i^\top\gamma + \rho e_i}
{\sqrt{1-\rho^2}}\right)\right],\quad e_i = \tfrac{y_i - f(x_i)}{\sigma}.
\label{eq:ll}
\end{equation}
Surveys and critiques: \citet{vella1998,puhani2000,winship1992}. The
selection-on-observables special case ($\rho = 0$) reduces to missing at
random \citep{rubin1976,little2002}.

\paragraph{Covariate shift and importance weighting.} When selection
depends only on observables, reweighting by inverse propensities is
consistent \citep{shimodaira2000,sugiyama2007}; \citet{cortes2008} give
the sample-selection-bias learning theory in exactly this
importance-weighting framework. All of it conditions on ignorability
given $x$; none of it addresses $\rho \neq 0$.

\paragraph{Selective labels and PU learning.} The selective-labels
literature \citep{lakkaraju2017} shows that decisions hide outcomes and
proposes contraction-based \emph{evaluation}; it diagnoses the problem
rather than repairing the predictive distribution. Positive-unlabeled
learning \citep{elkan2008,bekker2020} concerns missing labels for a
binary class under (usually) selected-completely-at-random assumptions,
not the MNAR sampling of $(x, y)$ pairs.

\paragraph{Flexible-ML selection models.} Closest in spirit is a recent
line that fits Heckman/Tobit selection models with flexible ML rather than
linear equations: \citet{oneill2025tobart} give a Type-2 Tobit sample-
selection model with Bayesian additive regression trees, a Dirichlet-
process-mixture error model that relaxes bivariate normality, and
individual credible intervals. It is the nearest prior work and we
distinguish it precisely: it is a tree ensemble (not a deep net), targets
econometric estimation and treatment effects (not epistemic uncertainty
under distribution shift), and does not study the selected-vs-well-sampled
region split or the importance-weighting failure that is our headline. Our
contribution is orthogonal --- selection-corrected \emph{epistemic UQ} with
deep feature maps --- and complementary: their non-normal error model is a
natural way to relax our Gaussian assumption (\S\ref{sec:e2}).

\paragraph{Deep uncertainty.} Deep ensembles
\citep{lakshminarayanan2017}, MC dropout \citep{gal2016}, GPs
\citep{rasmussen2006}, and quantile-calibration metrics
\citep{kuleshov2018} are the baselines and measurement tools we use;
\citet{ovadia2019} evaluate them under covariate shift. To our knowledge
--- after a kill-test against a 3M-abstract index and an adversarial web
sweep (2026-07-04, re-swept before submission; \S\ref{sec:repro}) --- no
deep-learning instantiation of selection-corrected epistemic UQ exists;
the nearest lines are those above plus the flexible-ML Tobit work just
cited, and we claim exactly the gap between them: \emph{when the sampling
of training points is correlated with unobserved determinants of the
outcome, importance weighting cannot fix the bias even with oracle
propensities; a joint selection/outcome error model can, given an
instrument.}

\section{Method: Heckman-corrected predictive distributions}
\label{sec:method}

\paragraph{Architecture.} The outcome function $f_\theta$ is a deep MLP;
the selection index $g_\psi(w) = w^\top\gamma + b$ is linear in $w = [x,
z]$ (a deliberately shallow selection head; the instrument $z$ enters
here and only here). Two global parameters $(\log\sigma,
\mathrm{atanh}\,\rho)$ complete the model.

\paragraph{Joint MLE.} We minimize $-\ell$ of Eq.~\eqref{eq:ll} over all
units --- unselected units contribute $\log\Phi(-g_\psi)$ --- with Adam.
With a flexible $f_\theta$ the correction term is at risk of being
absorbed into the outcome function early in training (the likelihood is
then locally flat in $\rho$); freezing $\rho = 0$ for a short warm-up
(100 of 1500 epochs), so that $f_\theta$ first fits the selected-sample
conditional mean, resolves this: post-warm-up, the $z$-variation of the
Mills term separates the correction from $f_\theta$. This is the
pre-registered methods contingency (\S\ref{sec:e1}); without warm-up
tuning, $\hat\rho$ collapses toward $0$.

\paragraph{Two-step.} A classic probit on $w$ over all units gives
$\hat\gamma$; the network then fits $y \approx f_\theta(x) + c\,
\IMR(w^\top\hat\gamma)$ on selected units by Gaussian NLL with $c$ a free
scalar, and $\sigma$ is recovered by the classic residual correction.
The population prediction is $f_\theta(x)$ alone.

\paragraph{Corrected predictive distribution.} An ensemble of $K = 5$
independently initialized models gives
\begin{equation}
\hat y(x) = \tfrac1K\sum_k f_{\theta_k}(x), \qquad
\widehat{\mathrm{Var}}(x) = \underbrace{\mathrm{Var}_k[f_{\theta_k}(x)]}_{\text{epistemic}}
+ \underbrace{\tfrac1K\sum_k \hat\sigma_k^2}_{\text{aleatoric}},
\end{equation}
i.e., exactly the deep-ensemble recipe with the population function in
place of the selected-sample one. Baselines (\S\ref{sec:e1}) get the same
ensemble treatment.

\section{A-E0: the faithfulness gate}
\label{sec:e0}

House rule: no experiment runs until the estimators reproduce classic
econometric results. On the \citet{cameron2005} RAND HIE specification
($n = \EZeroN$, \EZeroNSel{} selected; no exclusion restriction ---
their example of identification by functional form), our two-step matches
their published seven-digit Stata output with maximum absolute deviation
\EZeroDevTS{} across all coefficients, $\rho$, $\sigma$, and
$\beta_\IMR$; our MLE matches with maximum deviation \EZeroDevMLE{}
(log-likelihood difference \EZeroLLDev{}; $\hat\rho = \EZeroRhoMLE$,
$\hat\sigma = \EZeroSigmaMLE$). The probit head agrees with statsmodels
to $10^{-8}$; a statsmodels-composed two-step on the \citet{mroz1987}
data (Greene's example 22.8 spec) is matched to $10^{-8}$; and on
synthetic data with known truth ($n = 10^5$, $\rho = 0.6$) both
estimators recover $(\beta, \rho, \sigma)$ while naive OLS on the
selected sample does not, with the correct $\rho = 0$ MAR control.
Eleven tests, all green (\texttt{tests/}).

\section{A-E1: controlled demonstration}
\label{sec:e1}

\begin{figure}[t]
\centering
\includegraphics[width=\linewidth]{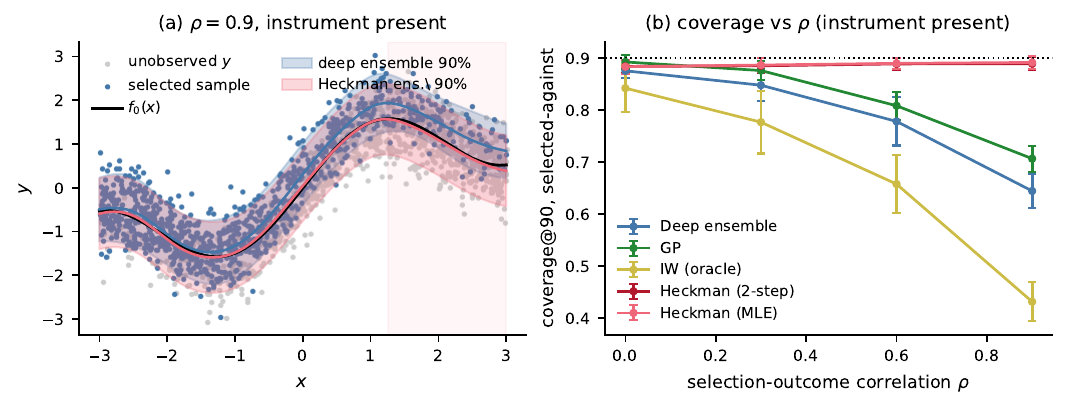}
\caption{(a) One draw at $\rho = 0.9$: baselines fit the selected sample
faithfully and are therefore biased with tight intervals in the
selected-against region (shaded); the Heckman ensemble recovers $f_0$.
(b) Coverage of nominal-90\% intervals in selected-against regions vs
$\rho$ (instrument present, 8 seeds, mean$\pm$sd).}
\label{fig:setup}
\end{figure}

\begin{figure}[t]
\centering
\includegraphics[width=\linewidth]{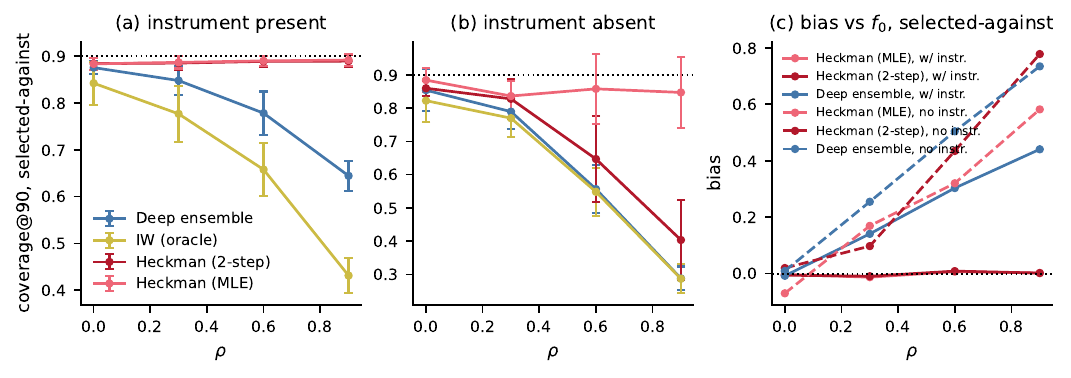}
\caption{The honesty curves. (a) With instrument ($\alpha = 1$) coverage
in selected-against regions is restored by the Heckman ensembles.
(b) Without the instrument ($\alpha = 0$) identification rides on
functional form; the MLE's coverage looks deceptively high but only
because weak identification inflates its intervals --- panel (c) shows the
mean is still badly biased. (c) Bias of the predictive mean against the
true $f_0$ in selected-against regions (solid: with instrument; dashed:
without): with the instrument the Heckman bias is $\approx 0$; without it,
all methods including Heckman stay biased.}
\label{fig:honesty}
\end{figure}

\paragraph{Generator.} $x \sim U[-3,3]$; instrument $z \sim N(0,1)$;
$(\varepsilon, u)$ bivariate normal with $\mathrm{Corr} = \rho$, $\sigma
= 0.5$; $y = \sin(1.5x) + 0.5x + \varepsilon$; selection index $g_0(x) +
\alpha z + u$ with $g_0$ decreasing in $x$, so high-$x$ regions are
smoothly selected against ($P(s{=}1\mid x)$ spans $\approx 0.05$--$0.95$;
overall \EOneSelFrac{} selected). Grid: $\rho \in \{0, 0.3, 0.6, 0.9\}
\times \alpha \in \{0, 1\} \times 8$ seeds; $n = 2000$ training pool;
metrics on a fresh uniform test set against population draws, split by
region ($P(s{=}1 \mid x) \le 0.3$ is ``selected-against''). Methods: deep
ensembles \citep{lakshminarayanan2017}, MC dropout \citep{gal2016}, an
RBF GP \citep{rasmussen2006}, an importance-weighted ensemble with
\emph{oracle} propensities $P(s{=}1 \mid x, z)$ from the generator, a
selection-blind two-head ablation (the Heckman architecture with $\rho
\equiv 0$: is it the joint error model or just the extra head?), and the
two Heckman ensembles. Figure~\ref{fig:setup} shows the setup and the
headline curves; Figure~\ref{fig:honesty} the instrument-present/absent
contrast.

\paragraph{(i) Baselines are miscalibrated where it matters.} At $\rho =
0.9$ with instrument present, deep-ensemble coverage in selected-against
regions is \CovAgEnsHi{} (nominal 90\%; $\rho = 0.6$: \CovAgEnsMid{}),
with predictive-mean bias \BiasAgEnsHi{} in $y$-units ($\sigma = 0.5$) ---
all members converge to the same biased $\E[y \mid x, s{=}1]$, so
ensemble spread cannot signal the error. MC dropout and the GP behave
alike (Fig.~\ref{fig:setup}b). At $\rho = 0$ (MAR control) everything is
fine --- the failure is specifically selection on unobservables, not
sparsity.

\paragraph{(ii) Oracle importance weighting does not fix it.} The IW
ensemble receives the exact generator propensities $P(s{=}1 \mid x,z)$
and still covers only \CovAgIWHi{} (bias \BiasAgIWHi{}) --- worse than
the unweighted ensemble, since upweighting the sparse selected-against
points concentrates training on units whose noise is most strongly
selected. This is the pre-registered kill condition and it does not
fire: IW with oracle propensities does not match the Heckman correction
under selection on unobservables (paired Wilcoxon on selected-against
coverage, $p = \PKillIWvsHeck$, $n = 8$ seeds; the Wilcoxon floor at
$n = 8$ is $p = 0.008$).

\paragraph{(iii) The Heckman correction does.} With the instrument, the
two-step ensemble covers \CovAgHeckTSHi{} at $\rho = 0.9$
(\CovAgHeckTSMid{} at $0.6$) with bias \BiasAgHeckTSHi{} --- an order of
magnitude better calibrated than the baselines in region-ECE
(\ECEAgHeckTSHi{} vs the deep ensemble's \ECEAgEnsHi{}, approaching the
oracle's \ECEAgOracleHi{}). The joint-MLE ensemble recovers the
correlation ($\hat\rho = \RhoHatHi \pm \RhoHatHiSD$ at true $0.9$) and
covers \CovAgHeckHi{}, matching the two-step here. The blind two-head
ablation stays at \CovAgBlindHi{} --- the gain is the joint error model,
not the extra head.

\paragraph{(iv) Without the instrument.} At $\alpha = 0$ the exclusion
restriction is gone and identification rides entirely on the
bivariate-normal form. Point estimates stay badly biased for every method,
Heckman included: at $\rho = 0.9$ the MLE's selected-against mean bias is
\BiasAgHeckHiNoInstr{} and the two-step's \BiasAgHeckTSHiNoInstr{}, versus
\BiasAgHeckTSHi{} \emph{with} the instrument, and $\hat\rho$ collapses to
\RhoHatHiNoInstr{} (true $0.9$). The interval coverage tells a
deceptive-looking story that we flag rather than exploit: the MLE's
selected-against coverage is \emph{higher} without the instrument
(\CovAgHeckHiNoInstr{}) than the two-step's (\CovAgHeckTSHiNoInstr{}), but
only because weak identification scatters its ensemble members and inflates
the predictive variance --- it covers by being \emph{uncertain}, not by
being right, as the large residual bias shows. The two-step, which lacks
that variance cushion, exposes the damage directly
(\CovAgHeckTSHiNoInstr{} vs \CovAgEnsHiNoInstr{} for the ensemble). The
lesson is the classic one \citep{puhani2000}, sharpened: without an
instrument, neither the point estimate nor the coverage should be trusted,
and coverage can actively mislead. Practitioners should locate themselves
on Figure~\ref{fig:honesty} before trusting any correction.

\paragraph{Methods finding: two-step vs joint MLE.} The joint MLE with
deep feature maps is stable \emph{only with the warm-up schedule}: frozen
$\rho = 0$ for the first 100 epochs, the outcome net first fits the
selected-sample mean and the Mills term then separates cleanly, giving
$\hat\rho = \RhoHatHi \pm \RhoHatHiSD$ (true $0.9$) and coverage
\CovAgHeckHi{} that matches the two-step. Without warm-up, $f_\theta$
absorbs the correction and $\hat\rho$ collapses toward $0$ (we observed
$\hat\rho \to 0.16$ at true $0.8$ in tuning). The two-step needs no such
schedule --- its hard separation of the probit fit from the outcome fit
is inherently robust --- so we recommend it as the default and report both
throughout.

\section{A-E2/E3: semi-real tabular MNAR}
\label{sec:e2}

We induce MNAR selection on two real regression datasets --- California
housing \citep{pace1997} and UCI wine quality \citep{cortez2009} --- by a
fully documented rule (code in \texttt{heckesel/synth.py}): the selection
index combines the standardized linear prediction of $y$ (so
high-predicted regions are selected against, loan-style), the
\emph{residual} of that fit scaled by $\rho$ (the unobservable), an
instrument $z \sim N(0,1)$, and a threshold set for 50\% selection.
Because we induce the selection, ground truth exists for all rows;
held-out test rows are never selected on. Grid: $\rho \in \{0, 0.5,
0.8\}$; 8 seeds at the headline $\rho = 0.8$, 3 elsewhere (house rule);
the same seven methods plus a \emph{skyline} ensemble trained on the
complete pre-selection pool (an oracle upper reference).

\begin{figure}[t]
\centering
\includegraphics[width=\linewidth]{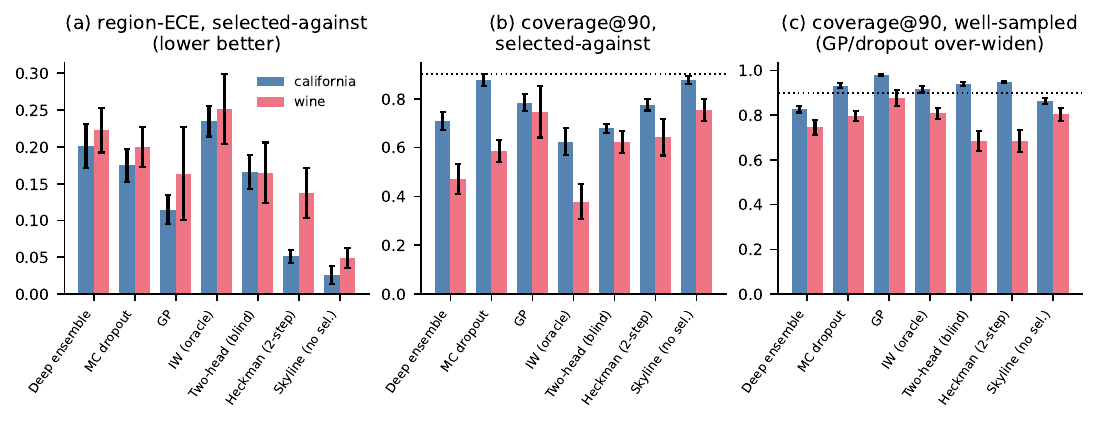}
\caption{Semi-real MNAR ($\rho = 0.8$, mean$\pm$sd over 8 seeds).
(a) Region-ECE in selected-against regions --- the calibration metric ---
where the two-step Heckman is the best non-oracle method on both datasets.
(b) Selected-against coverage is muddier: conservative baselines can match
it. (c) But those baselines get there by over-widening \emph{everywhere}
--- the GP covers \RealCovWellCalGP{} of a nominal-90\% interval in
well-sampled California regions --- which region-ECE penalizes and
coverage alone does not.}
\label{fig:real}
\end{figure}

Real heteroscedastic data makes the coverage metric treacherous, so we
lead with region-ECE (calibration error by region), the metric that
cannot be gamed by global over-widening. At $\rho = 0.8$ the corrected
two-step ensemble has the lowest selected-against ECE of every non-oracle
method on \emph{both} datasets --- \RealECECalHeckTS{} (California) and
\RealECEWineHeckTS{} (wine), versus the deep ensemble's \RealECECalEns{} /
\RealECEWineEns{}, oracle-IW's \RealECECalIW{} / \RealECEWineIW{}, MC
dropout's \RealECECalDrop{} / \RealECEWineDrop{}, and the GP's
\RealECECalGP{} / \RealECEWineGP{} --- approaching the skyline's
\RealECECalSky{} / \RealECEWineSky{} (paired one-sided Wilcoxon, two-step
below each of deep-ensemble, IW, and dropout, pooled over datasets:
$p < 0.001$ in all three). By raw \emph{coverage} the picture is muddier
(Fig.~\ref{fig:real}b): MC dropout and the GP sometimes match or exceed
the two-step's selected-against coverage (\RealCovCalHeckTS{} /
\RealCovWineHeckTS{}) --- but they do so by widening intervals
indiscriminately, over-covering the \emph{well-sampled} regions
(\RealCovWellCalGP{} / \RealCovWellCalDrop{} on California at nominal
90\%; Fig.~\ref{fig:real}c) rather than by locating and correcting the
selection bias. Oracle-IW remains the worst-calibrated method in
selected-against regions on both datasets, as the theory predicts. The
correction is not free of cost --- even the skyline undercovers wine
(non-Gaussian noise), and no method reaches nominal coverage on these
hard tasks --- but it is the only one that targets the selection bias
directly rather than blunting all its intervals. At $\rho = 0$ the
selected-against coverages coincide (\RealCovZeroCalEns{} vs
\RealCovZeroCalHeckTS{} on California), so the correction costs nothing
under MAR. As a naturally selected companion, the RAND HIE fit of
\S\ref{sec:e0} estimates $\hat\rho = \EZeroRhoMLE$ on real
medical-expenditure data --- with the caveat, stated by
\citet{cameron2005} themselves, that without an instrument this rests on
functional form; we make no ground-truth claim there.

\section{A-E4: benchmark reporting panels (a vignette)}
\label{sec:e4}

Public model$\times$task benchmark matrices are
missing-not-at-random panels: a paper reports its method on a chosen
subset of standard datasets. On the frozen Papers-with-Code evaluation
dump \citep{pwcdump2021} we build, per task, a panel over the eight
most-reported datasets (modal accuracy-like metric per dataset,
$z$-scored among reported values) and fit the classic Heckman MLE:
outcome $=$ cell $z$-score on a leave-one-out ability proxy and dataset
dummies; selection $=$ reported indicator on the same covariates. There
is \emph{no} exclusion restriction here --- precisely the regime our
controlled experiments warn about --- and the fits behave accordingly:
\VigNBoundary{} of \VigNTotal{} panel and pairwise fits hit the
$|\hat\rho| = 1$ boundary, the textbook no-instrument pathology. The
interior estimates are suggestive --- ImageNet$\to$ImageNet-ReaL
reporting selects positively on unobservables ($\hat\rho =
\VigImageNetReaLRho \pm \VigImageNetReaLSE$) --- but sign-unstable across
panels (machine translation: $\VigMTRho$). We present this as a
methodological caution, not an estimate: \emph{benchmark-panel
corrections need a reporting instrument} (venue page limits and dataset
release dates are candidates), and until then, leaderboard aggregations
inherit an unidentified selection bias.

\begin{figure}[t]
\centering
\includegraphics[width=0.62\linewidth]{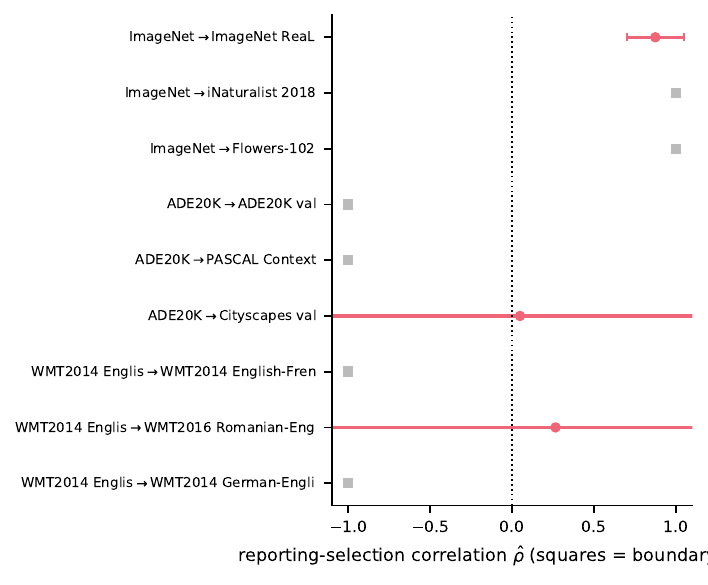}
\caption{Heckman $\hat\rho$ on Papers-with-Code reporting panels
(anchor$\to$target pairs). Squares: fits on the $|\hat\rho| = 1$
boundary. Without an instrument, most fits degenerate --- the real-world
face of Figure~\ref{fig:honesty}b.}
\label{fig:vignette}
\end{figure}

\section{What we do not claim}

We make no causal-discovery claims: the correction assumes the
two-equation structure, it does not find it. We make no real-MNAR
ground-truth claims beyond the induced-selection experiments; on RAND HIE
and the reporting panels the truth is unknowable without instruments, and
we say so. We do not claim the bivariate-normal error model is correct
for any real dataset --- only that, where its identification conditions
hold even approximately, it beats reweighting approaches that are
structurally unable to address the problem. We do not claim superiority
under selection on observables ($\rho = 0$), where importance weighting
is consistent and simpler.

\section{Limitations}

The outcome noise is modeled as homoscedastic Gaussian; real data
(\S\ref{sec:e2}) already shows the cost, as even the skyline undercovers.
Heteroscedastic and non-Gaussian extensions (e.g., $t$-copula errors) are
natural next steps. The selection head is linear; a nonlinear selection
index with a valid instrument is compatible with the likelihood but
untested here. Ensembles of size 5 approximate epistemic variance
coarsely. The vignette panels aggregate heterogeneous metrics and papers;
they are illustrative only.

\section{Reproducibility}
\label{sec:repro}

All code, tests, committed result JSONs, and this manuscript's
auto-generated numbers (\texttt{numbers.tex}; regenerate-and-diff passes
at submission) are in the repository; every number in the paper is
machine-generated from \texttt{results/*.json} (discipline inherited from
\citealp{howe2026eventrice}). Experiments run on one RTX 3080 (10\,GB) in
under six GPU-hours total. Datasets: RandHIE and Mroz87 from the
sampleSelection R package's public mirrors \citep{toomet2008}; the
Cameron--Trivedi Stata reference output is committed alongside. Archived
release: \href{https://doi.org/10.5281/zenodo.21210466}{DOI:
10.5281/zenodo.21210466}. Pre-submission recency sweep logged in
\texttt{PLAN.md}.

\bibliographystyle{plainnat}
\bibliography{references}

\end{document}